\def\year{2019}\relax
\documentclass[letterpaper]{article} 
\usepackage{aaai19}  
\usepackage{times}  
\usepackage{helvet}  
\usepackage{courier}  
\usepackage{url}  
\usepackage{graphicx}  
\usepackage{tipa}
\usepackage{booktabs}
\usepackage{amsmath}

\graphicspath{{./fig/}}

\newcommand{\citet}[1]
{\citeauthor{#1}~\shortcite{#1}}
\newcommand{\citep}{\cite}

\title{Neural Speech Synthesis with Transformer Network}

\author{
Naihan Li\thanks{Work done during internship at Microsoft STC Asia.}\textsuperscript{1,4},
Shujie Liu\textsuperscript{2}, Yanqing Liu\textsuperscript{3}, Sheng Zhao\textsuperscript{3},
Ming Liu\textsuperscript{1,4}, Ming Zhou\textsuperscript{2}\\
\textsuperscript{1}{University of Electronic Science and Technology of China}\\
\textsuperscript{2}{Microsoft Research Asia}\\
\textsuperscript{3}{Microsoft STC Asia}\\
\textsuperscript{4}{CETC Big Data Research Institute Co.,Ltd, Guizhou, China}\\
\url{lnhzsbls1994@163.com}\\
\url{{shujliu, yanqliu, szhao, mingzhou}@microsoft.com}\\
\url{csmliu@uestc.edu.cn}\\
}

\begin{document}
%
\maketitle
\begin{abstract}
Although end-to-end neural text-to-speech (TTS) methods (such as Tacotron2) are proposed and achieve state-of-the-art performance, they still suffer from two problems: 1) low efficiency during training and inference; 2) hard to model long dependency using current recurrent neural networks (RNNs).
Inspired by the success of Transformer network in neural machine translation (NMT), in this paper, we introduce and adapt the multi-head attention mechanism to replace the RNN structures and also the original attention mechanism in Tacotron2. With the help of multi-head self-attention, the hidden states in the encoder and decoder are constructed in parallel, which improves training efficiency. Meanwhile, any two inputs at different times are connected directly by a self-attention mechanism, which solves the long range dependency problem effectively.
Using phoneme sequences as input, our Transformer TTS network generates mel spectrograms, followed by a WaveNet vocoder to output the final audio results. Experiments are conducted to test the efficiency and performance of our new network. For the efficiency, our Transformer TTS network can speed up the training about 4.25 times faster compared with Tacotron2. For the performance, rigorous human tests show that our proposed model achieves state-of-the-art performance (outperforms Tacotron2 with a gap of 0.048) and is very close to human quality (4.39 vs 4.44 in MOS).
\end{abstract}

\section{Introduction}

Text to speech (TTS) is a very important task for user interaction, aiming to synthesize intelligible and natural audios which are indistinguishable from human recordings. Traditional TTS systems have two components: front-end and back-end. Front-end is responsible for text analysis and linguistic feature extraction, such as word segmentation, part of speech tagging, multi-word disambiguation and prosodic structure prediction; back-end is built for speech synthesis based on linguistic features from front-end, such as speech acoustic parameter modeling, prosody modeling and speech generation. In the past decades, concatenative and parametric speech synthesis systems were mainstream techniques. However, both of them have complex pipelines, and defining good linguistic features is often time-consuming and language specific, which requires a lot of resource and manpower. Besides, synthesized audios often have glitches or instability in prosody and pronunciation compared to human speech, and thus sound unnatural.

Recently, with the rapid development of neural networks, end-to-end generative text-to-speech models, such as Tacotron \cite{wang2017tacotron} and Tacotron2 \cite{shen2017natural}, are proposed to simplify traditional speech synthesis pipeline by replacing the production of these linguistic and acoustic features with a single neural network. Tacotron and Tacotron2 first generate mel spectrograms directly from texts, then synthesize the audio results by a vocoder such as Griffin Lim algorithm \cite{griffin1984signal} or WaveNet \cite{van2016wavenet}. With the end-to-end neural network, quality of synthesized audios is greatly improved and even comparable with human recordings on some datasets. The end-to-end neural TTS models contain two components, an encoder and a decoder. Given the input sequence (of words or phonemes), the encoder tries to map them into a semantic space and generates a sequence of encoder hidden states, and the decoder, taking these hidden states as context information with an attention mechanism, constructs the decoder hidden states then outputs the mel frames. For both encoder and decoder, recurrent neural networks (RNNs) are usually leveraged, such as LSTM \cite{hochreiter1997long} and GRU \cite{Cho2014Learning}.

However, RNNs can only consume the input and generate the output sequentially, since the previous hidden state and the current input are both required to build the current hidden state. The characteristic of sequential process limits the parallelization capability in both the training and inference process. For the same reason, for a certain frame, information from many steps ahead may has been biased after multiple recurrent processing. To deal with these two problems, Transformer \cite{vaswani2017attention} is proposed to replace the RNNs in NMT models.

Inspired by this idea, in this paper, we combine the advantages of Tacotron2 and Transformer to propose a novel end-to-end TTS model, in which the multi-head attention mechanism is introduced to replace the RNN structures in the encoder and decoder, as well as the vanilla attention network. The self-attention mechanism unties the sequential dependency on the last previous hidden state to improve the parallelization capability and relieve the long distance dependency problem. Compared with the vanilla attention between the encoder and decoder, the multi-head attention can build the context vector from different aspects using different attention heads. With the phoneme sequences as input, our novel Transformer TTS network generates mel spectrograms, and employs WaveNet as vocoder to synthesize audios. We conduct experiments with 25-hour professional speech dataset, and the audio quality is evaluated by human testers. Evaluation results show that our proposed model outperforms the original Tacotron2 with a gap of 0.048 in CMOS, and achieves a similar performance (4.39 in MOS) with human recording (4.44 in MOS).  Besides, our Transformer TTS model can speed up the training process about 4.25 times compared with Tacotron2.
Audio samples can be accessed on \url{https://neuraltts.github.io/transformertts/}

\section{Background}
In this section, we first introduce the sequence-to-sequence model, followed by a brief description about Tacotron2 and Transformer, which are two preliminaries in our work.
\subsection{Sequence to Sequence Model}

A sequence-to-sequence model \cite{sutskever2014sequence,bahdanau2014neural} converts an input sequence $(x_1,x_2,...,x_T)$ into an output sequence $(y_1,y_2,...,y_{T'})$, and each predicted $y_t$ is conditioned on all previously predicted outputs $y_1,...,y_{t-1}$. In most cases, these two sequences are of different lengths ($T \neq T'$).
In NMT, this conversion translates the input sentence in one language into the output sentence in another language, based on a conditional probability $p(y_1,...,y_T'|x_1,...,x_T)$:
\begin{align}
h_t &= encoder(h_{t-1}, x_t) \\
s_t &= decoder(s_{t-1}, y_{t-1}, c_t)
\end{align}
where $c_t$ is the context vector calculated by an attention mechanism:
\begin{align}
c_t &= attention(s_{t-1}, \mathbf{h})
\end{align}
thus $p(y_1,...,y_T'|x_1,...,x_T)$ can be computed by
\begin{equation}
p(y_1,...,y_T'|x_1,...,x_T)=\prod_{t=1}^{T'} p(y_t|\mathbf{y_{<t}},\mathbf{x})
\end{equation}
and
\begin{equation}
p(y_t|\mathbf{y_{<t}},\mathbf{x}) = softmax(f(s_t))
\end{equation}
where $f(\cdot)$ is a fully connected layer. For translation tasks, this softmax function is among all dimensions of $f(s_t)$ and calculates the probability of each word in the vocabulary. However, in the TTS task, the softmax function is not required and the hidden states $\mathbf{s}$ calculated by decoder are consumed directly by a linear projection to  obtain the desired spectrogram frames.

\subsection{Tacotron2}

\begin{figure}[tb]
  \centering
  \includegraphics[width=0.85\linewidth]{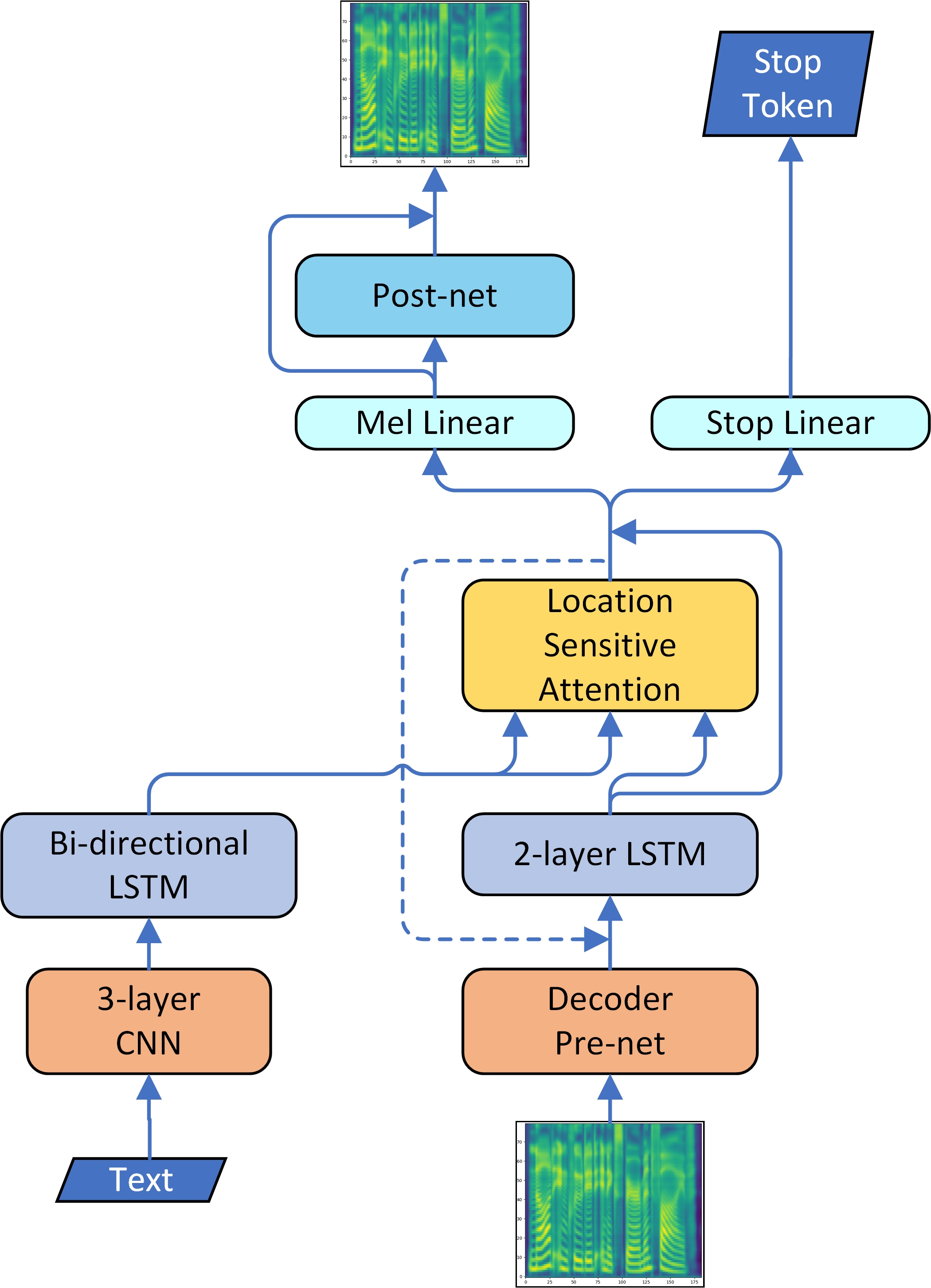}
  \caption{System architecture of Tacotron2.}
  \label{fig:tacotron2}
\end{figure}

Tacotron2 is a neural network architecture for speech synthesis directly from text, as shown in Fig. \ref{fig:tacotron2} . The embedding sequence of input is firstly processed with a 3-layer CNN to extract a longer-term context, and then fed into the encoder, which is a bi-directional LSTM. The previous mel spectrogram frame (the predicted one in inference, or the golden one in training time), is first processed with a 2-layer fully connected network (decoder pre-net), whose output is concatenated with the previous context vector, followed by a 2-layer LSTM. The output is used to calculate the new context vector at this time step, which is concatenated with the output of the 2-layer LSTM to predict the mel spectrogram and stop token with two different linear projections respectively. Finally the predicted mel spectrogram is fed into a 5-layer CNN with residual connections to refine the mel spectrogram.

\subsection{Transformer for NMT}
\label{subsec:transformer}

\begin{figure}[!tb]
  \centering
  \includegraphics[width=0.98\linewidth]{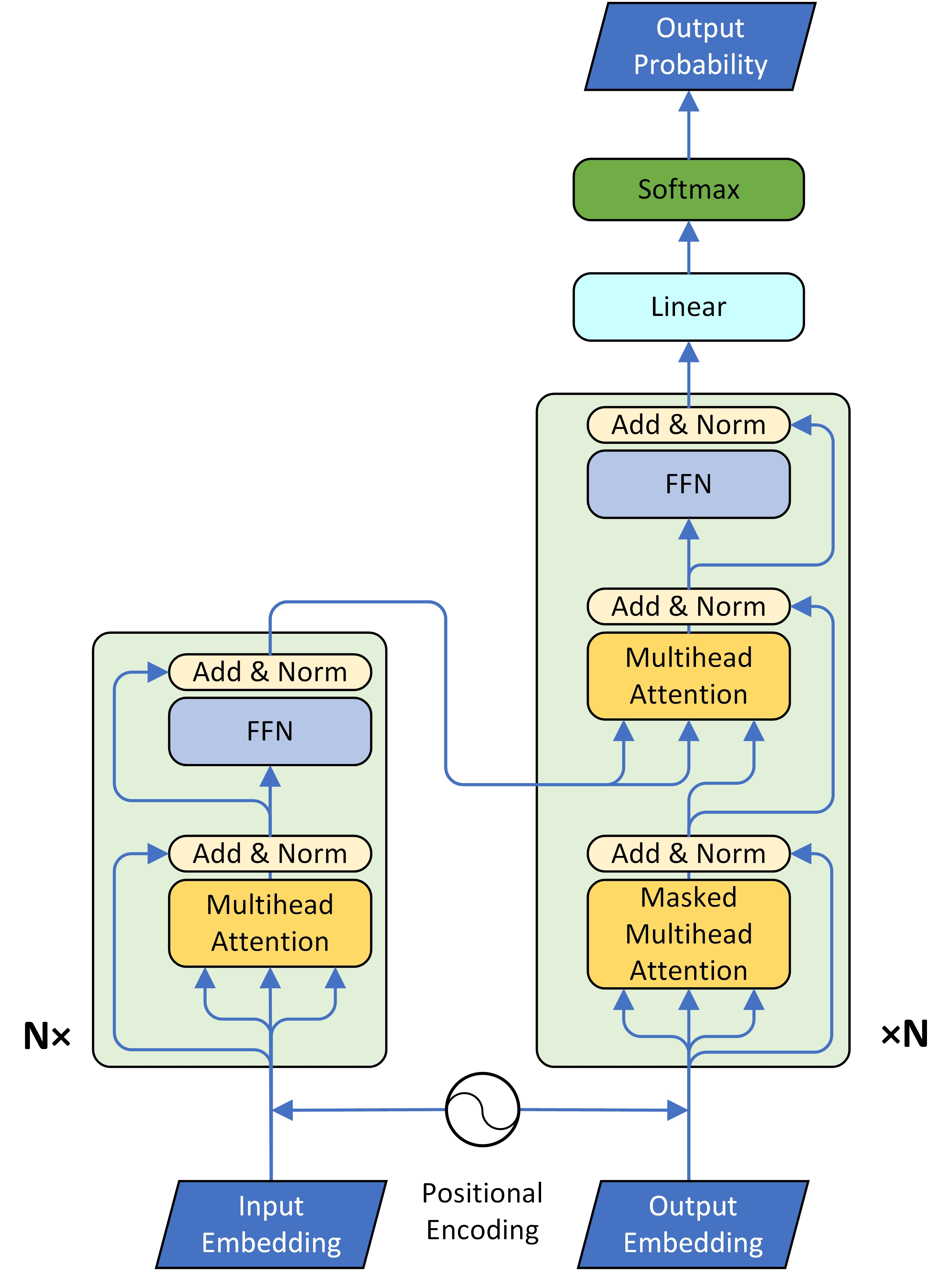}
  \caption{System architecture of Transformer.}
  \label{fig:transformer}
\end{figure}

Transformer \cite{vaswani2017attention}, shown in Fig. \ref{fig:transformer}, is a sequence to sequence network, based solely on attention mechanisms and dispensing with recurrences and convolutions entirely. In recent works, Transformer has shown extraordinary results, which outperforms many RNN-based models in NMT. It consists of two components: an encoder and a decoder, both are built by stacks of several identity blocks. Each encoder block contains two subnetworks: a multi-head attention and a feed forward network, while each decoder block contains an extra masked multi-head attention comparing to the encoder block. Both encoder and decoder blocks have residual connections and layer normalizations.

\section{Neural TTS with Transformer}

Compared to RNN-based models, using Transformer in neural TTS has two advantages. First it enables parallel training by removing recurrent connections, as frames of an input sequence for decoder can be provided in parallel. The second one is that self attention provides an opportunity for injecting global context of the whole sequence into each input frame, building long range dependencies directly. Transformer shortens the length of paths forward and backward signals have to traverse between any combination of positions in the input and output sequences down to 1. This helps a lot in a neural TTS model, such as the prosody of synthesized waves, which not only depends on several words in the neighborhood, but also sentence level semantics.

\begin{figure}[!htb]
  \centering
  \includegraphics[width=0.97\linewidth]{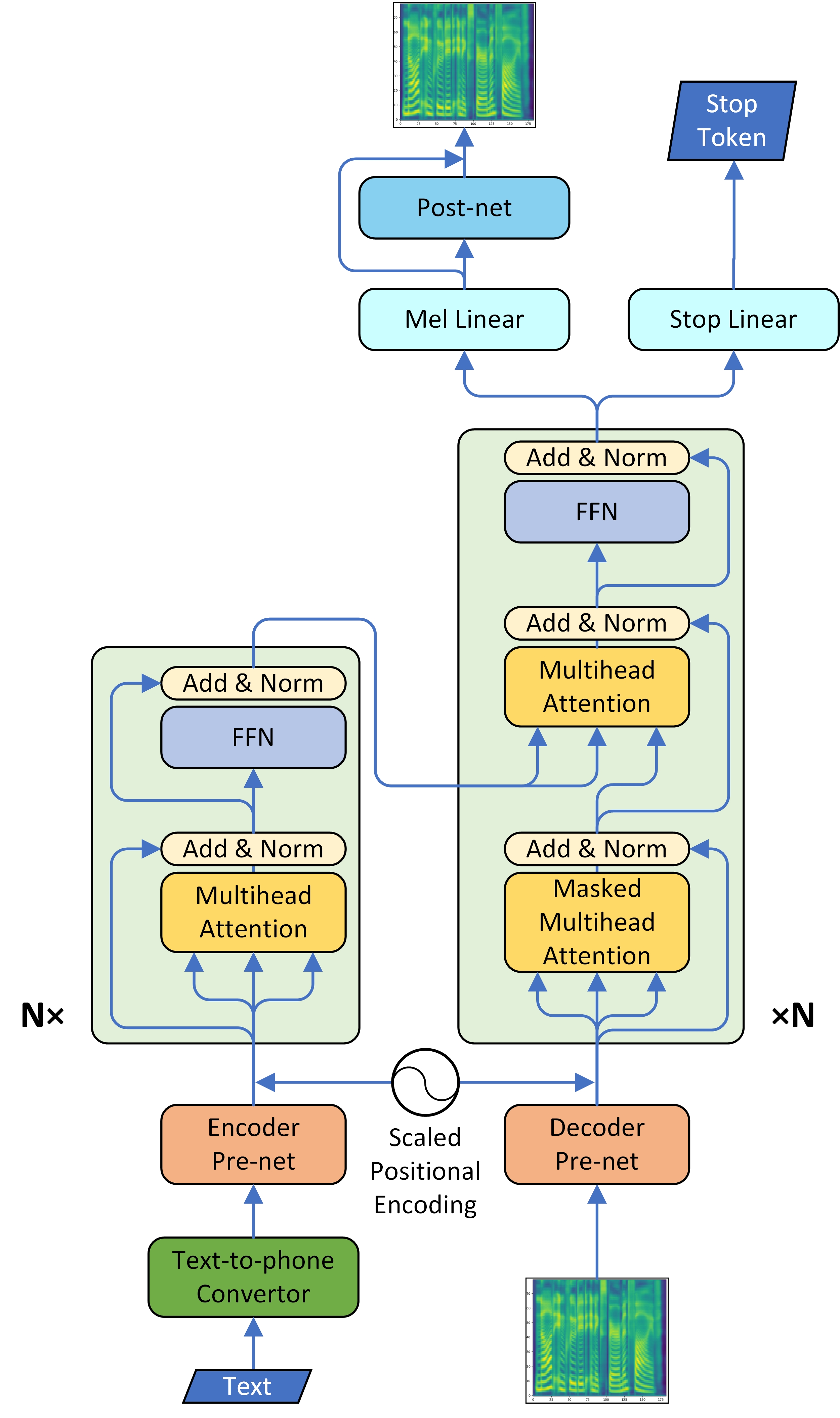}
  \caption{System architecture of our model.}
  \label{fig:transtron}
\end{figure}

In this section we will introduce the architecture of our Transformer TTS model, and analyze the function of each part. The overall structure diagram is shown in Fig. \ref{fig:transtron}.

\subsection{Text-to-Phoneme Converter}

English pronunciation has certain regularities, for example, there are two kinds of syllables in English: open and closed. The letter "a" is often pronounced as /e\i/ when it's in an open syllable, while it is pronounced as /\ae/ or /a\textlengthmark/ in closed syllables. We can rely on the neural network to learn such a regularity in the training process. However, it is difficult to learn all the regularities when, which is often the case, the training data is not sufficient enough, and some exceptions have too few occurrences for neural networks to learn. So we make a rule system and implement it as a text-to-phoneme converter, which can cover the vast majority of cases.

\subsection{Scaled Positional Encoding}
\label{subsec:scl_tri_pe}

Transformer contains no recurrence and no convolution so that if we shuffle the input sequence of encoder or decoder, we will get the same output. To take the order of the sequence into consideration, information about the relative or absolute position of frames is injected by triangle positional embeddings, shown in Eq. \ref{eq:tri_pe}:
\begin{align}
  PE(pos,2i)&=\sin(\frac{pos}{10000^{\frac{2i}{d_{model}}}}) \\
  PE(pos,2i+1)&=\cos(\frac{pos}{10000^{\frac{2i}{d_{model}}}})
  \label{eq:tri_pe}
\end{align}
where $pos$ is the time step index, $2i$ and $2i+1$ is the channel index and $d_{model}$ is the vector dimension of each frame. In NMT, the embeddings for both source and target language are from language spaces, so the scales of these embeddings are similar. This condition doesn't hold in the TTS scenarioe, since the source domain is of texts while the target domain is of mel spectrograms, hence using fixed positional embeddings may impose heavy constraints on both the encoder and decoder pre-nets (which will be described in Sec. \ref{subsec:enc_prenet} and \ref{subsec:dec_prenet}). We employ these triangle positional embeddings with a trainable weight, so that these embedding can adaptively fit the scales of both encoder and decoder pre-nets' output, as shown in Eq. \ref{eq:scl_tri_pe}:
\begin{equation}
  x_i = prenet({phoneme}_i)+\alpha PE(i)
  \label{eq:scl_tri_pe}
\end{equation}
where $\alpha$ is the trainable weight.

\subsection{Encoder Pre-net}
\label{subsec:enc_prenet}
In Tacotron2, a 3-layer CNN is applied to the input text embeddings, which can model the longer-term context in the input character sequence. In our Transformer TTS model, we input the phoneme sequence into the same network, which is called "encoder pre-net". Each phoneme has a trainable embedding of 512 dims, and the output of each convolution layer has 512 channels, followed by a batch normalization and ReLU activation, and a dropout layer as well. In addition, we add a linear projection after the final ReLU activation, since the output range of ReLU is $[0,+\infty)$, while each dimension of these triangle positional embeddings is in $[-1,1]$. Adding 0-centered positional information onto non-negative embeddings will result in a fluctuation not centered on the origin and harm model performance, which will be demonstrated in our experiment. Hence we add a linear projection for center consistency.

\subsection{Decoder Pre-net}
\label{subsec:dec_prenet}
The mel spectrogram is first consumed by a neural network composed of two fully connected layers(each has 256 hidden units) with ReLU activation, named "decoder pre-net", and it plays an important role in the TTS system.
Phonemes has trainable embeddings thus their subspace is adaptive, while that of mel spectrograms is fixed. We infer that decoder pre-net is responsible for projecting mel spectrograms into the same subspace as phoneme embeddings, so that the similarity of a $\left \langle phoneme, mel\ frame \right \rangle$ pair can be measured, thus the attention mechanism can work. Besides, 2 fully connected layers without non-linear activation are also tried but no reasonable attention matrix aligning the hidden states of encoder and decoder can be generated. In our other experiment, hidden size is enlarged from 256 to 512, however that doesn't generate significant improvement but needs more steps to converge. Accordingly, we conjecture that mel spectrograms have a compact and low dimensional subspace that 256 hidden units are good enough to fit. This conjecture can also be evidenced in our experiment, which is shown in Sec. \ref{subsubsec:pe_method}, that the final positional embedding scale of decoder is smaller than that of encoder. An additional linear projection is also added like encoder pre-net not only for center consistency but also obtain the same dimension as the triangle positional embeddings.

\subsection{Encoder}
\label{subsec:encoder}
In Tacotron2, the encoder is a bi-directional RNN. We replace it with Transformer encoder which is described in Sec. \ref{subsec:transformer} . Comparing to original bi-directional RNN, multi-head attention splits one attention into several subspaces so that it can model the frame relationship in multiple different aspects, and it directly builds the long-time dependency between any two frames thus each of them considers global context of the whole sequence. This is crucial for synthesized audio prosody especially when the sentence is long, as generated samples sound more smooth and natural in our experiments. In addition, employing multi-head attention instead of original bi-directional RNN can enable parallel computing to improve training speed.

\subsection{Decoder}

In Tacotron2, the decoder is a 2-layer RNN with location-sensitive attention \cite{chorowski2015attention}. We replace it with Transformer decoder which is described in Sec. \ref{subsec:transformer}.
Employing Transformer decoder makes two main differences, adding self-attention, which can bring similar advantages described in Sec. \ref{subsec:encoder}, and using multi-head attention instead of the location-sensitive attention. The multi-head attention can integrate the encoder hidden states in multiple perspectives and generate better context vectors.
Taking attention matrix of previous decoder time steps into consideration, location-sensitive attention used in Tacotron2 can encourage the model to generate consistent attention results.
We try to modify the dot product based multi-head attention to be location sensitive, but that doubles the training time and easily run out of memory.

\subsection{Mel Linear, Stop Linear and Post-net}
Same as Tacotron2, we use two different linear projections to predict the mel spectrogram and the stop token respectively, and use a 5-layer CNN to produce a residual to refine the reconstruction of mel spectrogram. It's worth mentioning that, for the stop linear, there is only one positive sample in the end of each sequence which means "stop", while hundreds of negative samples for other frames. This imbalance may result in unstoppable inference. We impose a positive weight ($5 .0\sim 8.0$) on the tail positive stop token when calculating binary cross entropy loss, and this problem was efficiently solved.

\section{Experiment}

In this section, we conduct experiments to test our proposed Transformer TTS model with 25-hour professional speech pairs, and the audio quality is evaluated by human testers in MOS and CMOS.

\subsection{Training Setup}

We use 4 Nvidia Tesla P100 to train our model with an internal US English female dataset, which contains 25-hour professional speech (17584 $\left \langle text, wave \right \rangle$ pairs, with a few too long waves removed). 50ms silence at head and 100ms silence at tail are kept for each wave. Since the lengths of training samples vary greatly, fixed batch size will either run out of memory when long samples are added into a batch with a large size or waste the parallel computing power if the batch is small and into which short samples are divided. Therefore, we use the dynamic batch size where the maximum total number of mel spectrogram frames is fixed and one batch should contain as many samples as possible. Thus there are on average 16 samples in single batch per GPU. We try training on a single GPU, but the procedures are quiet instable or even failed, by which synthesized audios were like raving and incomprehensible. Even if training doesn't fail, synthesized waves are of bad quality and weird prosody, or even have some severe problems like missing phonemes. Thus we enable multi-GPU training to enlarge the batch size, which effectively solves those problems.

\subsection{Text-to-Phoneme Conversion and Pre-process}

Tacotron2 uses character sequences as input, while our model is trained on pre-normalized phoneme sequences. Word and syllable boundaries, punctuations are also included as special markers. The process pipeline to get training phoneme sequences contains sentence separation, text normalization, word segmentation and finally obtaining pronunciation. By text-to-phoneme conversion, mispronunciation problems are greatly reduced especially for those pronunciations that are rarely occurred in our training set.

\subsection{WaveNet Settings}

We train a WaveNet conditioned on mel spectrogram with the same internal US English female dataset, and use it as the vocoder for all models in this paper.
The sample rate of ground truth audios is 16000 and frame rate (frames per second) of ground truth mel spectrogram is 80.
Our autoregressive WaveNet contains 2 QRNN layers and 20 dilated layers, and the sizes of all residual channels and dilation channels are all 256. Each frame of QRNN's final output is copied 200 times to have the same spatial resolution as audio samples and be conditions of 20 dilated layers.

\subsection{Training Time Comparison}
Our model can be trained in parallel since there is no recurrent connection between frames. In our experiment, time consume in a single training step for our model is $\sim$0.4s, which is 4.25 times faster than that of Tacotron2 ($\sim$1.7s) with equal batch size (16 samples per batch). However, since the parameter quantity of our model is almost twice than Tacotron2, it still takes $\sim$3 days to converge comparing to $\sim$4.5 days of that for Tacotron2.

\subsection{Evaluation}

We randomly select 38 fixed examples with various lengths (no overlap with training set) from our internal dataset as the evaluation set. We evaluate mean option score (MOS) on these 38 sentences generated by different models (include recordings), in which case we can keep the text content consistent and exclude other interference factors hence only examine audio quality.
For higher result accuracy, we split the whole MOS test into several small tests, each containing one group from our best model, one group from a comparative model and one group of recordings. Those MOS tests are rigorous and reliable, as \textbf{each audio is listened to by at least 20 testers, who are all native English speakers } (comparing to Tacotron2's 8 testers in \citet{shen2017natural}), and each tester listens less than 30 audios.

We train a Tacotron2 model with our internal US English female dataset as the baseline (also use phonemes as input), and gain equal MOS with our model. Therefore we test the comparison mean option score (CMOS) between samples generated by Tacotron2 and our model for a finer contrast. In the comparison mean option score (CMOS) test, testers listen to two audios (generated by Tacotron2 and our model with the same text) each time and evaluates how the latter feels comparing to the former using a score in $\left[-3, 3\right]$ with intervals of 1. The order of the two audios changes randomly so testers don't know their sources. Our model wins by a gap of 0.048, and detailed results are shown in Table \ref{tab:mos_system}.

\begin{table}[t!]
  \centering
  \begin{tabular}{ccc}
    \toprule
    System     &   MOS &   CMOS  \\
    \midrule
    Tacotron2           &   $4.39 \pm 0.05$ &   $0$\\
    Our Model           &   $4.39 \pm 0.05$ &   \textbf{0.048}\\
    \midrule
    Ground Truth        &   $4.44 \pm 0.05$ &   -\\
    \bottomrule
  \end{tabular}
  \caption{MOS comparison among our model, our Tacotron2 and recordings.}
  \label{tab:mos_system}
\end{table}

\begin{figure}[t!]
  \centering
  \includegraphics[width=0.95\linewidth]{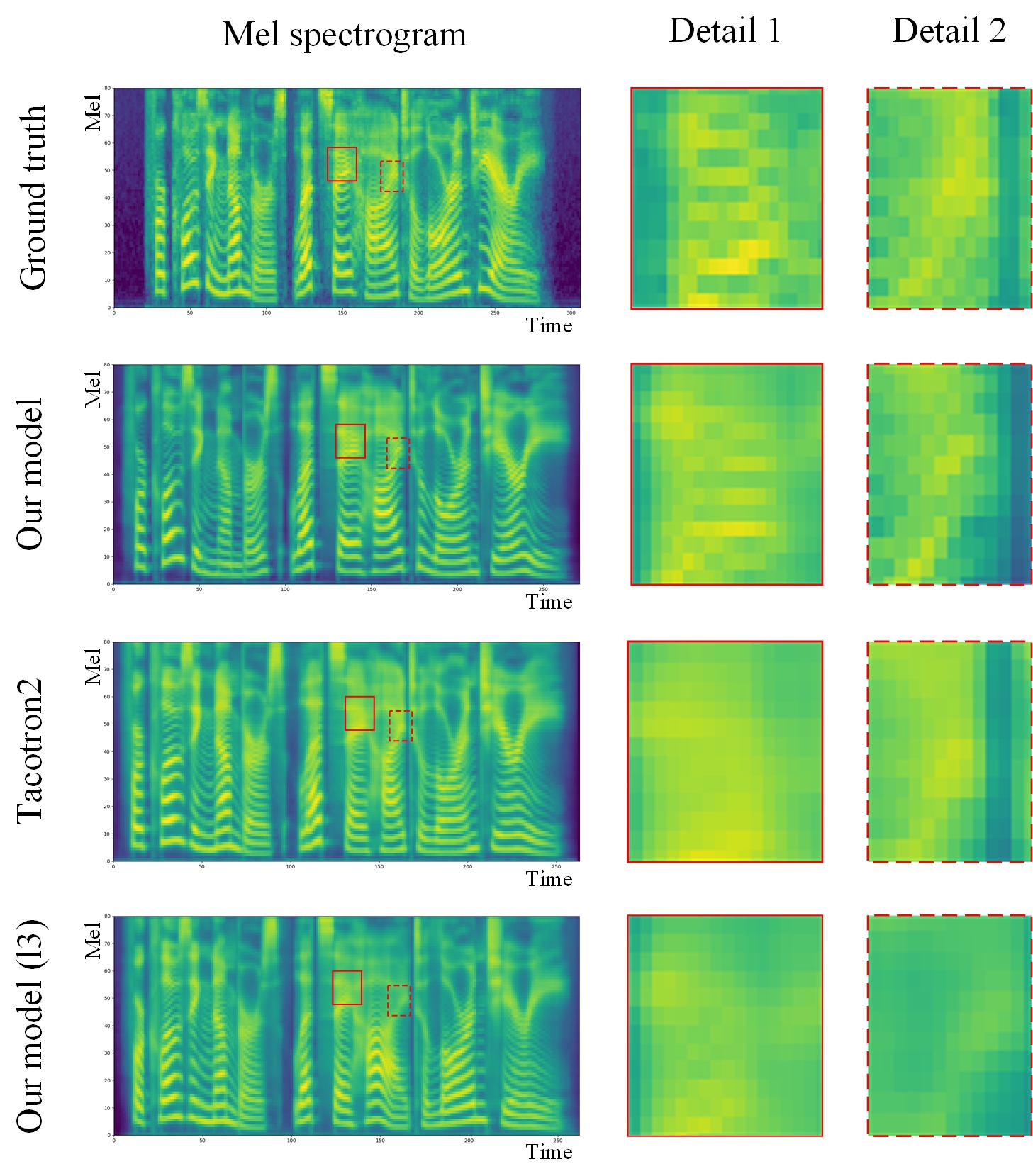}
  \caption{Mel spectrogram comparison. Our model (6-layer) does better in reconstructing details as marked in red rectangles, while Tacotron2 and our 3-layer model blur the texture especially in high frequency region. Best viewed in color.}
  \label{fig:mel_comp}
\end{figure}

We also select mel spectrograms generated by our model and Tacotron2 respectively with the same text, and compare them together with ground truth, as shown in column 1,2 and 3 of Fig. \ref{fig:mel_comp}. As we can see, our model does better in reconstructing details as marked in red rectangles, while Tacotron2 left out the detailed texture in high frequency region.

\begin{figure}[t!]
  \centering
  \includegraphics[width=0.95\linewidth]{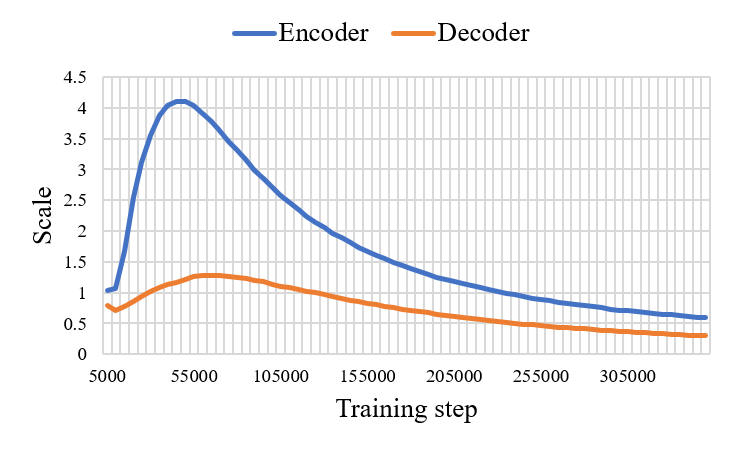}
  \caption{PE scale of encoder and decoder.}
  \label{fig:pe_scale}
\end{figure}

\subsection{Ablation Studies}

In this section, we study the detail modification of network architecture, and conduct several experiments to show our improvements.

\subsubsection{Re-centering Pre-net's Output}

As described in Sec. \ref{subsec:enc_prenet} and \ref{subsec:dec_prenet}, we re-project both the encoder and decoder pre-nets' outputs for consistent center with positional embeddings. In contrast, we add no linear projection in encoder pre-net and add a fully connected layer with ReLU activation in decoder pre-net. The results imply that center-consistent positional embedding performs slightly better, as shown in Table \ref{tab:mos_re_projection}.

\begin{table}[t!]
  \centering
  \begin{tabular}{ccc}
    \toprule
    Re-center    &   MOS    \\
    \midrule
    No            &   $4.32 \pm 0.05$               \\
    Yes     &   \textbf{4.36} $\pm 0.05$           \\
    \midrule
    Ground Truth        &   $4.43 \pm 0.05$        \\
    \bottomrule
  \end{tabular}
	\caption{MOS comparison of whether re-centering pre-net's output.}
  \label{tab:mos_re_projection}
\end{table}

\subsubsection{Different Positional Encoding Methods}
\label{subsubsec:pe_method}

We inject positional information into both encoder's and decoder's input sequences as Eq. \ref{eq:scl_tri_pe}. Fig. \ref{fig:pe_scale} shows that the final positional embedding scales of encoder and decoder are different, and Table \ref{tab:mos_scale_type} shows model with trainable scale performs slightly better. We think that the trainable scale relaxes the constraint on encoder and decoder pre-nets, making positional information more adaptive for different embedding spaces.

We also try adding absolute position embeddings (each position has a trainable embedding) to the sequence, which also works but has some severe problems such as missing phonemes when the sequences became long. That's because long sample is relatively rare in the training set, so the embeddings for large indexes can hardly be trained and thus the position information won't be accurate for rear frames in a long sample.

\subsubsection{Model with Different Hyper-Parameter}

Both the encoder and decoder of the original Transformer is composed of 6 layers, and each multi-head attention has 8 heads. We compare performance and training speed with different layer and head numbers, as shown in Table \ref{tab:mos_ln}, \ref{tab:mos_hn} and \ref{tab:speed_hp}. We find that reducing layers and heads both improve the training speed, but on the other hand, harm model performance in different degrees.

We notice that in both the 3-layer and 6-layer model, only alignments from certain heads of the beginning 2 layers' are interpretable diagonal lines, which shows the approximate correspondence between input and output sequence, while those of the following layers are disorganized. Even so, more layers can still lower the loss, refine the synthesized mel spectrogram and improve audio quality. The reason is that with residual connection between different layers, our model fits target transformation in a Taylor-expansion way: the starting terms account most as low ordering ones, while the subsequential ones can refine the function. Hence adding more layer makes the synthesized wave more natural, since it does better in processing spectrogram details (shown in column 4, Fig. \ref{fig:mel_comp}). Fewer heads can slightly reduce training time cost since there are less production per layer, but also harm the performance.

\begin{table}[t!]
  \centering
  \begin{tabular}{ccc}
    \toprule
    PE Type    &   MOS    \\
    \midrule
    Original            &   $4.37 \pm 0.05$               \\
    Scaled     &   \textbf{4.40} $\pm 0.05$           \\
    \midrule
    Ground Truth        &   $4.41 \pm 0.04$        \\
    \bottomrule
  \end{tabular}
  \caption{MOS comparison of scaled and original PE.}
  \label{tab:mos_scale_type}
\end{table}

\begin{table}[t!]
  \centering
  \begin{tabular}{ccc}
    \toprule
    Layer Number   &   MOS    \\
    \midrule
    3-layer                 &   $4.33 \pm 0.06$               \\
    6-layer        &   \textbf{4.41} $\pm 0.05$           \\
    \midrule
    Ground Truth            &   $4.44 \pm 0.05$        \\
    \bottomrule
  \end{tabular}
  \caption{Ablation studies in different layer numbers.}
  \label{tab:mos_ln}
\end{table}

\begin{table}[t!]
  \centering
  \begin{tabular}{ccc}
    \toprule
    Head Number   &   MOS    \\
    \midrule
    4-head                 &   $4.39 \pm 0.05$               \\
    8-head        &   \textbf{4.44} $\pm 0.05$           \\
    \midrule
    Ground Truth           &   $4.47 \pm 0.05$        \\
    \bottomrule
  \end{tabular}
  \caption{Ablation studies in different head numbers.}
  \label{tab:mos_hn}
\end{table}

\begin{table}[t!]
  \centering
  \begin{tabular}{c|cc}
    \toprule
    $ $                &   3-layer    &   6-layer  \\
    \hline
    4-head    &   -                   &   $0.44$~~~         \\
    8-head    &   $0.29$              &   $0.50$~~~         \\
    \bottomrule
  \end{tabular}
  \caption{Comparison of time consuming (in second) per training step of different layer and head numbers.}
  \label{tab:speed_hp}
\end{table}

\section{Related Work}

Traditional speech synthesis methods can be categorized into two classes: concatenative systems and parametric systems. Concatenative TTS systems \cite{hunt1996unit,black1997automatically} split original waves into small units, and stitch them by some algorithms such as Viterbi \cite{viterbi1967error} followed by signal process methods \cite{charpentier1986diphone,verhelst1993overlap} to generate new waves. Parametric TTS systems \cite{tokuda2000speech,zen2009statistical,ze2013statistical,tokuda2013speech} convert speech waves into spectrograms, and acoustic parameters,  such as fundamental frequency and duration, are used to synthesize new audio results.

Traditional speech synthesis methods require extensive domain expertise and may contain brittle design choices. Char2Wav \cite{sotelo2017char2wav} integrates the front-end and the back-end as one seq2seq \cite{sutskever2014sequence,bahdanau2014neural} model and learns the whole process in an end-to-end way, predicting acoustic parameters followed by a SampleRNN \cite{mehri2016samplernn} as the vocoder. However, acoustic parameters are still intermediate for audios, thus Char2Wav is not a really end-to-end TTS model, and  their seq2seq and SampleRNN models need to be separately pre-trained,
while Tacotron, proposed by \citet{wang2017tacotron}, is an end-to-end generative text-to-speech model, which can be trained by  $\left \langle text, spectrogram \right \rangle$ pairs directly from scratch, and synthesizes speech audios with generated spectrograms by Griffin Lim algorithm \cite{griffin1984signal}.
Based on Tacotron, Tacotron2 \cite{shen2017natural}, a unified and entirely neural model, generates mel spectrograms by a Tacotron-style neural network and then synthesizes speech audios by a modified WaveNet \cite{van2016wavenet}. WaveNet is an autoregressive generative model for waveform synthesis, composed of stacks of dilated convolutional layers and processes raw audios of very high temporal resolution (e.g., 24,000 sample rate), while suffering from very large time cost in inference. This problem is solved by Parallel WaveNet \cite{oord2017parallel}, based on the inverse autoregressive flow (IAF) \cite{kingma2016improved} and reaches $1000 \times$ real time. Recently, ClariNet \cite{ping2018clarinet}, a fully convolutional text-to-wave neural architecture, is proposed to enable the fast end-to-end training from scratch.
Moreover, VoiceLoop \cite{taigman2018voiceloop} is an alternative neural TTS method mimicking a person's voice based on samples captured in-the-wild, such as audios of public speeches, and even with an inaccurate automatic transcripts.

On the other hand, Transformer \cite{vaswani2017attention} is proposed for neural machine translation (NMT) and achieves state-of-the-art result. Previous NMT models are dominated by RNN-based \cite{bahdanau2014neural} or CNN-based (e.g. ConvS2S \cite{gehring2017convolutional}, ByteNet \cite{kalchbrenner2016neural}) neural networks. For RNN-based models, both training and inference are sequential for each sample, while CNN-based models enable parallel training. Both RNN and CNN based models are difficult to learn dependencies between distant positions since RNNs have to traverse a long path and CNN has to stack many convolutional layers to get a large receptive field, while Transformer solves this using self attention in both its encoder and decoder. The ability of self-attention is also proved in SAGAN \cite{zhang2018self}, where original GANs without self-attention fail to capture geometric or structural patterns that occur consistently in some classes (for example, dogs are often drawn without clearly defined separate feet). By adding self-attention, these failure cases are greatly reduced.
Besides, multi-head attention is proposed to obtain different relations in multi-subspaces. Recently, Transformer has been applied in automatic speech recognition (ASR) \cite{zhou2018comparison,zhou2018syllable}, proving its ability in acoustic modeling other than natural language process.

\section{Conclusion and Future Work}

We propose a neural TTS model based on Tacotron2 and Transformer, and make some modification to adapt Transformer to neural TTS task. Our model generates audio samples of which quality is very closed to human recording, and enables parallel training and learning long-distance dependency so that the training is sped up and the audio prosody is much more smooth. We find that batch size is crucial for training stability, and more layers can refine the detail of generated mel spectrograms especially for high frequency regions thus improve model performance.

Even thought Transformer has enabled parallel training, autoregressive model still suffers from two problems, which are slow inference and exploration bias. Slow inference is due to the dependency of previous frames when infer current frame, so that the inference is sequential, while exploration bias comes from the autoregressive error accumulation. We may solve them both at once by building a non-autoregressive model, which is also our current research in progress.

\bibliography{mybib}

\begin{thebibliography}{}

\bibitem[\protect\citeauthoryear{Bahdanau, Cho, and
  Bengio}{2014}]{bahdanau2014neural}
Bahdanau, D.; Cho, K.; and Bengio, Y.
\newblock 2014.
\newblock Neural machine translation by jointly learning to align and
  translate.
\newblock {\em arXiv preprint arXiv:1409.0473}.

\bibitem[\protect\citeauthoryear{Black and
  Taylor}{1997}]{black1997automatically}
Black, A.~W., and Taylor, P.
\newblock 1997.
\newblock Automatically clustering similar units for unit selection in speech
  synthesis.
\newblock In {\em Fifth European Conference on Speech Communication and
  Technology}.

\bibitem[\protect\citeauthoryear{Charpentier and
  Stella}{1986}]{charpentier1986diphone}
Charpentier, F., and Stella, M.
\newblock 1986.
\newblock Diphone synthesis using an overlap-add technique for speech waveforms
  concatenation.
\newblock In {\em Acoustics, Speech, and Signal Processing, IEEE International
  Conference on ICASSP'86.}, volume~11,  2015--2018.
\newblock IEEE.

\bibitem[\protect\citeauthoryear{Cho \bgroup et al\mbox.\egroup
  }{2014}]{Cho2014Learning}
Cho, K.; Van~Merrienboer, B.; Gulcehre, C.; Bahdanau, D.; Bougares, F.;
  Schwenk, H.; and Bengio, Y.
\newblock 2014.
\newblock Learning phrase representations using rnn encoder-decoder for
  statistical machine translation.
\newblock {\em Computer Science}.

\bibitem[\protect\citeauthoryear{Chorowski \bgroup et al\mbox.\egroup
  }{2015}]{chorowski2015attention}
Chorowski, J.~K.; Bahdanau, D.; Serdyuk, D.; Cho, K.; and Bengio, Y.
\newblock 2015.
\newblock Attention-based models for speech recognition.
\newblock In {\em Advances in neural information processing systems},
  577--585.

\bibitem[\protect\citeauthoryear{Gehring \bgroup et al\mbox.\egroup
  }{2017}]{gehring2017convolutional}
Gehring, J.; Auli, M.; Grangier, D.; Yarats, D.; and Dauphin, Y.~N.
\newblock 2017.
\newblock Convolutional sequence to sequence learning.
\newblock {\em arXiv preprint arXiv:1705.03122}.

\bibitem[\protect\citeauthoryear{Griffin and Lim}{1984}]{griffin1984signal}
Griffin, D., and Lim, J.
\newblock 1984.
\newblock Signal estimation from modified short-time fourier transform.
\newblock {\em IEEE Transactions on Acoustics, Speech, and Signal Processing}
  32(2):236--243.

\bibitem[\protect\citeauthoryear{Hochreiter and
  Schmidhuber}{1997}]{hochreiter1997long}
Hochreiter, S., and Schmidhuber, J.
\newblock 1997.
\newblock Long short-term memory.
\newblock {\em Neural computation} 9(8):1735--1780.

\bibitem[\protect\citeauthoryear{Hunt and Black}{1996}]{hunt1996unit}
Hunt, A.~J., and Black, A.~W.
\newblock 1996.
\newblock Unit selection in a concatenative speech synthesis system using a
  large speech database.
\newblock In {\em Acoustics, Speech, and Signal Processing, 1996. ICASSP-96.
  Conference Proceedings., 1996 IEEE International Conference on}, volume~1,
  373--376.
\newblock IEEE.

\bibitem[\protect\citeauthoryear{Kalchbrenner \bgroup et al\mbox.\egroup
  }{2016}]{kalchbrenner2016neural}
Kalchbrenner, N.; Espeholt, L.; Simonyan, K.; Oord, A. v.~d.; Graves, A.; and
  Kavukcuoglu, K.
\newblock 2016.
\newblock Neural machine translation in linear time.
\newblock {\em arXiv preprint arXiv:1610.10099}.

\bibitem[\protect\citeauthoryear{Kingma \bgroup et al\mbox.\egroup
  }{2016}]{kingma2016improved}
Kingma, D.~P.; Salimans, T.; Jozefowicz, R.; Chen, X.; Sutskever, I.; and
  Welling, M.
\newblock 2016.
\newblock Improved variational inference with inverse autoregressive flow.
\newblock In {\em Advances in Neural Information Processing Systems},
  4743--4751.

\bibitem[\protect\citeauthoryear{Mehri \bgroup et al\mbox.\egroup
  }{2016}]{mehri2016samplernn}
Mehri, S.; Kumar, K.; Gulrajani, I.; Kumar, R.; Jain, S.; Sotelo, J.;
  Courville, A.; and Bengio, Y.
\newblock 2016.
\newblock Samplernn: An unconditional end-to-end neural audio generation model.
\newblock {\em arXiv preprint arXiv:1612.07837}.

\bibitem[\protect\citeauthoryear{Oord \bgroup et al\mbox.\egroup
  }{2017}]{oord2017parallel}
Oord, A. v.~d.; Li, Y.; Babuschkin, I.; Simonyan, K.; Vinyals, O.; Kavukcuoglu,
  K.; Driessche, G. v.~d.; Lockhart, E.; Cobo, L.~C.; Stimberg, F.; et~al.
\newblock 2017.
\newblock Parallel wavenet: Fast high-fidelity speech synthesis.
\newblock {\em arXiv preprint arXiv:1711.10433}.

\bibitem[\protect\citeauthoryear{Ping, Peng, and Chen}{2018}]{ping2018clarinet}
Ping, W.; Peng, K.; and Chen, J.
\newblock 2018.
\newblock Clarinet: Parallel wave generation in end-to-end text-to-speech.
\newblock {\em arXiv preprint arXiv:1807.07281}.

\bibitem[\protect\citeauthoryear{Shen \bgroup et al\mbox.\egroup
  }{2017}]{shen2017natural}
Shen, J.; Pang, R.; Weiss, R.~J.; Schuster, M.; Jaitly, N.; Yang, Z.; Chen, Z.;
  Zhang, Y.; Wang, Y.; Skerry-Ryan, R.; et~al.
\newblock 2017.
\newblock Natural tts synthesis by conditioning wavenet on mel spectrogram
  predictions.
\newblock {\em arXiv preprint arXiv:1712.05884}.

\bibitem[\protect\citeauthoryear{Sotelo \bgroup et al\mbox.\egroup
  }{2017}]{sotelo2017char2wav}
Sotelo, J.; Mehri, S.; Kumar, K.; Santos, J.~F.; Kastner, K.; Courville, A.;
  and Bengio, Y.
\newblock 2017.
\newblock Char2wav: End-to-end speech synthesis.
\newblock {\em ICLR 2017 workshop}.

\bibitem[\protect\citeauthoryear{Sutskever, Vinyals, and
  Le}{2014}]{sutskever2014sequence}
Sutskever, I.; Vinyals, O.; and Le, Q.~V.
\newblock 2014.
\newblock Sequence to sequence learning with neural networks.
\newblock In {\em Advances in neural information processing systems},
  3104--3112.

\bibitem[\protect\citeauthoryear{Taigman \bgroup et al\mbox.\egroup
  }{2018}]{taigman2018voiceloop}
Taigman, Y.; Wolf, L.; Polyak, A.; and Nachmani, E.
\newblock 2018.
\newblock Voiceloop: Voice fitting and synthesis via a phonological loop.
\newblock In {\em International Conference on Learning Representations}.

\bibitem[\protect\citeauthoryear{Tokuda \bgroup et al\mbox.\egroup
  }{2000}]{tokuda2000speech}
Tokuda, K.; Yoshimura, T.; Masuko, T.; Kobayashi, T.; and Kitamura, T.
\newblock 2000.
\newblock Speech parameter generation algorithms for hmm-based speech
  synthesis.
\newblock In {\em Acoustics, Speech, and Signal Processing, 2000. ICASSP'00.
  Proceedings. 2000 IEEE International Conference on}, volume~3,  1315--1318.
\newblock IEEE.

\bibitem[\protect\citeauthoryear{Tokuda \bgroup et al\mbox.\egroup
  }{2013}]{tokuda2013speech}
Tokuda, K.; Nankaku, Y.; Toda, T.; Zen, H.; Yamagishi, J.; and Oura, K.
\newblock 2013.
\newblock Speech synthesis based on hidden markov models.
\newblock {\em Proceedings of the IEEE} 101(5):1234--1252.

\bibitem[\protect\citeauthoryear{Van Den~Oord \bgroup et al\mbox.\egroup
  }{2016}]{van2016wavenet}
Van Den~Oord, A.; Dieleman, S.; Zen, H.; Simonyan, K.; Vinyals, O.; Graves, A.;
  Kalchbrenner, N.; Senior, A.~W.; and Kavukcuoglu, K.
\newblock 2016.
\newblock Wavenet: A generative model for raw audio.
\newblock In {\em SSW},  125.

\bibitem[\protect\citeauthoryear{Vaswani \bgroup et al\mbox.\egroup
  }{2017}]{vaswani2017attention}
Vaswani, A.; Shazeer, N.; Parmar, N.; Uszkoreit, J.; Jones, L.; Gomez, A.~N.;
  Kaiser, {\L}.; and Polosukhin, I.
\newblock 2017.
\newblock Attention is all you need.
\newblock In {\em Advances in Neural Information Processing Systems},
  5998--6008.

\bibitem[\protect\citeauthoryear{Verhelst and
  Roelands}{1993}]{verhelst1993overlap}
Verhelst, W., and Roelands, M.
\newblock 1993.
\newblock An overlap-add technique based on waveform similarity (wsola) for
  high quality time-scale modification of speech.
\newblock In {\em Acoustics, Speech, and Signal Processing, 1993. ICASSP-93.,
  1993 IEEE International Conference on}, volume~2,  554--557.
\newblock IEEE.

\bibitem[\protect\citeauthoryear{Viterbi}{1967}]{viterbi1967error}
Viterbi, A.
\newblock 1967.
\newblock Error bounds for convolutional codes and an asymptotically optimum
  decoding algorithm.
\newblock {\em IEEE transactions on Information Theory} 13(2):260--269.

\bibitem[\protect\citeauthoryear{Wang \bgroup et al\mbox.\egroup
  }{2017}]{wang2017tacotron}
Wang, Y.; Skerry-Ryan, R.; Stanton, D.; Wu, Y.; Weiss, R.~J.; Jaitly, N.; Yang,
  Z.; Xiao, Y.; Chen, Z.; Bengio, S.; et~al.
\newblock 2017.
\newblock Tacotron: A fully end-to-end text-to-speech synthesis model.
\newblock {\em arXiv preprint}.

\bibitem[\protect\citeauthoryear{Ze, Senior, and
  Schuster}{2013}]{ze2013statistical}
Ze, H.; Senior, A.; and Schuster, M.
\newblock 2013.
\newblock Statistical parametric speech synthesis using deep neural networks.
\newblock In {\em Acoustics, Speech and Signal Processing (ICASSP), 2013 IEEE
  International Conference on},  7962--7966.
\newblock IEEE.

\bibitem[\protect\citeauthoryear{Zen, Tokuda, and
  Black}{2009}]{zen2009statistical}
Zen, H.; Tokuda, K.; and Black, A.~W.
\newblock 2009.
\newblock Statistical parametric speech synthesis.
\newblock {\em Speech Communication} 51(11):1039--1064.

\bibitem[\protect\citeauthoryear{Zhang \bgroup et al\mbox.\egroup
  }{2018}]{zhang2018self}
Zhang, H.; Goodfellow, I.; Metaxas, D.; and Odena, A.
\newblock 2018.
\newblock Self-attention generative adversarial networks.
\newblock {\em arXiv preprint arXiv:1805.08318}.

\bibitem[\protect\citeauthoryear{Zhou \bgroup et al\mbox.\egroup
  }{2018a}]{zhou2018comparison}
Zhou, S.; Dong, L.; Xu, S.; and Xu, B.
\newblock 2018a.
\newblock A comparison of modeling units in sequence-to-sequence speech
  recognition with the transformer on mandarin chinese.
\newblock {\em arXiv preprint arXiv:1805.06239}.

\bibitem[\protect\citeauthoryear{Zhou \bgroup et al\mbox.\egroup
  }{2018b}]{zhou2018syllable}
Zhou, S.; Dong, L.; Xu, S.; and Xu, B.
\newblock 2018b.
\newblock Syllable-based sequence-to-sequence speech recognition with the
  transformer in mandarin chinese.
\newblock {\em arXiv preprint arXiv:1804.10752}.

\end{thebibliography}
\bibliographystyle{aaai}

\end{document}